# ROBOT VISION: CALIBRATION OF WIDE-ANGLE LENS CAMERAS USING COLLINEARITY CONDITION AND K-NEAREST NEIGHBOUR REGRESSION

J. C. K. Chow [1,2,3*], I. Detchev [4], K. D. Ang [3,5], K. Morin [6], K. Mahadevan [7], N. Louie [3]

[1] Department of Medicine, Cumming School of Medicine, University of Calgary, Calgary, Alberta, Canada – jckchow@ucalgary.ca
[2] School of Earth and Planetary Sciences, Faculty of Science and Engineering, Curtin University, Perth, WA, Australia
[3] Department of Research and Development, Vusion Technologies, Calgary, Alberta, Canada – niclouie@outlook.com
[4] Department of Geomatics Engineering, Schulich School of Engineering, University of Calgary, Calgary, Alberta, Canada – i.detchev@ucalgary.ca
[5] Department of Computer Science, Faculty of Science, University of Calgary, Calgary, Alberta, Canada – kdang@ucalgary.ca
[6] Leica Geosystems, Heerbrugg, Canton of St. Gallen, Switzerland – kristian.morin@leica-geosystems.com
[7] Department of Electrical and Computer Engineering, Faculty of Engineering, University of Alberta, Edmonton, Alberta, Canada – karthik@ualberta.ca

**Commission I, ICWG I/IV**

**ABSTRACT:**

Visual perception is regularly used by humans and robots for navigation. By either implicitly or explicitly mapping the environment, ego-motion can be determined and a path of actions can be planned. The process of mapping and navigation are delicately intertwined; therefore, improving one can often lead to an improvement of the other. Both processes are sensitive to the interior orientation parameters of the camera system and mathematically modelling these systematic errors can often improve the precision and accuracy of the overall solution. This paper presents an automatic camera calibration method suitable for any lens, without having prior knowledge about the sensor. Statistical inference is performed to map the environment and localize the camera simultaneously. K-nearest neighbour regression is used to model the geometric distortions of the images. A normal-angle lens Nikon camera and wide-angle lens GoPro camera were calibrated using the proposed method, as well as the conventional bundle adjustment with self-calibration method (for comparison). Results showed that the mapping error was reduced from an average of 14.9 mm to 1.2 mm (i.e. a 92% improvement) and 66.6 mm to 1.5 mm (i.e. a 98% improvement) using the proposed method for the Nikon and GoPro cameras, respectively. In contrast, the conventional approach achieved an average 3D error of 0.9 mm (i.e. 94% improvement) and 3.3 mm (i.e. 95% improvement) for the Nikon and GoPro cameras, respectively. Thus, the proposed method performs well irrespective of the lens/sensor used: it yields results that are comparable to the conventional approach for normal-angle lens cameras, and it has the additional benefit of improving calibration results for wide-angle lens cameras.

## 1. INTRODUCTION

An average human eye has a vertical field-of-view (FOV) of approximately 135 degrees (°) and a horizontal FOV of 160°. With binocular field the horizontal FOV extends to 200° and beyond. This implicitly assists in our day-to-day human activities such as navigation, path-planning, object recognition and tracking. Such a wide visual field is beneficial for survival in the animal kingdom because it allows more information to be gathered and analysed from a single viewpoint without exerting energy to turn our heads. For example, obstacles and predators in our peripheral vision would be undetectable if our visual field was narrower.

When building mobile robots, e.g. unmanned aerial vehicles (UAVs), self-driving cars, and humanoid robots, engineers often draw on inspiration from nature. For instance, the wide FOV of the human visual system is advantageous for autonomous navigation. In recent years, wide-angle lenses have gained popularity in robotics. Within the growing drone market, majority of UAVs are equipped with shorter focal length lens cameras. For example, the DJI Mavic PRO has a camera with 26 mm focal length, and the Phantom 4 has a camera with 20 mm focal length. Action cameras (e.g. GoPro) often have a FOV exceeding 140°. In fact, even dating back to 1970, Nikon demonstrated a camera with a 220° FOV at the Photokina exhibition that could take pictures of the environment behind the sensor.

While wide-angle lens cameras are beneficial in robot vision, the larger FOV introduces some new challenges that need to be addressed before these cameras can be used for ego-motion estimation and structure from motion. Firstly, conventional close-range photogrammetry often utilizes digital cameras equipped with a normal-angle lens. To illustrate, a typical 35 mm camera with a 50 mm focal length lens yields vertical and horizontal FOVs of approximately 27° and 40°, respectively. On the other hand, wide-angle lens cameras allow more light rays to enter with large incidence angles (i.e. significantly deviant from the optical axis), and consequently more distortions can be observed in the images. In fact, there is a singularity in the conventional collinearity equations when the incidence angle is perpendicular to the optical axis and bundle adjustment cannot be performed. Furthermore, the ground sampling distance grows as we move towards the periphery of the image. This results in significant variation in image measurement quality within the image plane due to changes in resolution.

The abovementioned challenges have been addressed by various authors. Some methods are limited to a FOV of less than 180° and others require an expert photogrammetrist to tune the parameters (Luhmann et al., 2006). In this article, a fully-automated calibration method suitable for central cameras with any FOV is presented. The proposed method is based on an alternative formulation of the collinearity equations, and all latent variables are learned automatically from the data. No expert knowledge is necessary since machine learning approaches are used to automatically adapt the tuning hyperparameters and decide on the model complexity based on the dataset.

## 2. BACKGROUND

In the world of wide-angle camera systems, there are a variety of operating principles. The most basic and perhaps well-known is the fisheye lens camera, which achieves a wider FOV based on the construction of the lens itself. To calibrate fisheye lens camera systems, Schwalbe (2005) presents a model that assumes a near-linear relationship between the incidence angle and the radial distance (measured from the optical axis) in image space. In this study, a calibration room with targets arranged in concentric circles was used and ground truth target locations were required since single photo resection was used rather than bundle adjustment. Additionally, since the mathematical model replaced the principal distance with radius of the image circle as the scale factor, prior knowledge about the lens' focal length cannot easily be incorporated. Furthermore, based on the fisheye projective model developed, any object ray with an incidence angle of 90° is undefined.

Another study (Schneider et al., 2009) further explored different geometric models for fisheye lenses, specifically looking at four different lens constructions (equidistance, equisolid-angle, stereographic, and orthographic projection) and their corresponding models. In this case, the models were tested using both spatial resection and bundle adjustment as calibration methods. For both calibration methods, a standard deviation of 1/10 of a pixel was achieved, with the bundle adjustment results being slightly worse than the resection results. Although four different models were tested, both the interior orientation parameters (IOP) and exterior orientation parameters (EOP) did not differ significantly between the different models. On the other hand, the additional parameters (AP) did vary significantly between models and were able to compensate for the differences between them. The study also suggests using image variant IOP, which yielded better image space precision but reduced object space accuracy. Furthermore, in their check point analysis, systematic trends were still observed: calibration took care of approximately 75% of systematic effects. Overall, the study demonstrated that using the appropriate lens model to match the lens construction yielded the best results but may not always be practical since the lens construction is not always known from the manufacturer's specifications. In these cases, calibration becomes labour-intensive as multiple models need to be tested to find the most suitable one.

Aside from fisheye lenses, there is another method for capturing omnidirectional images – catadioptric systems. These systems use some combination of mirror and conventional camera: for example, a parabolic mirror with an orthographic camera, or a hyperbolic or elliptical mirror with a perspective camera. Going beyond merely calibration of fisheye lens systems, Scaramuzza et al. (2006a) developed a sensor-independent method which can be used for calibration of both catadioptric systems and fisheye lens systems. A fourth-order polynomial was determined to be most suitable for the model. The calibration was tested using a catadioptric system with 200° FOV (KAIDAN 360° One VR hyperbolic mirror and a SONY CCD camera) and an accuracy of approximately one pixel was achieved. However, the method assumed that the imaging projection function is rotationally symmetric and required a visible circular external boundary, which is prohibitive to being used for full format fisheye lenses. Furthermore, based on their website, their MATLAB toolbox for calibrating omnidirectional cameras (Scaramuzza et al. 2006b) which implemented their sensor independent method only works with a planar target field and on cameras with a FOV of up to 195°. This FOV limitation can be attributed to using a Taylor polynomial, which is a function that does not allow mapping on a full sphere. Urban et al. (2015) later improved the convergence and efficacy of this calibration method by modifying the residual function to estimate the IOP and EOP simultaneously.

## 3. MATHEMATICAL MODEL

The modified bundle adjustment with self-calibration method proposed in this paper is inspired by Netter's anatomical drawing of the human visual field (Figure 1) and is grounded by the theory of photogrammetry. Anatomically, light from a person's environment intersects at a common point near the anterior aspect of the eye before hitting the cones distributed on a curved surface near the posterior aspect of the eye.

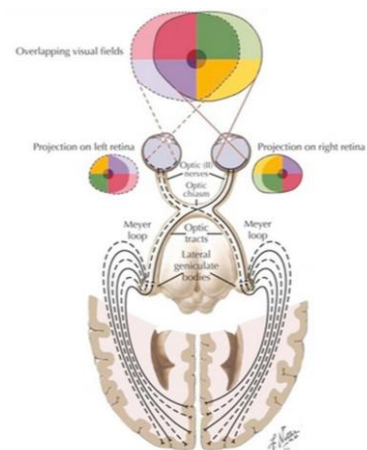

Figure 1: Frank H. Netter's illustration of the human visual field (Netter, 2014)

Based on the assumption in photogrammetry that light rays travel in a straight line over a short distance (i.e. in the absence of atmospheric refraction), any deviation in the direction of the light ray can be attributed to the AP of the camera system. In conventional photogrammetry with normal lenses, this is mostly caused by radial lens distortion (RLD) and decentering lens distortion (DLD). For wide-angle lenses with FOV less than 180° it may be sufficient in certain applications to model the effect of larger incidence angles by including higher-order terms in the classical polynomial RLD model. However, in the standard collinearity equations, a singularity exists when the incidence angle approaches 90°. In such cases, an object point will be projected into infinity in the image space. One solution is to model the lens as a curved surface rather than a pin-hole – after all, a fisheye lens is indeed a physically high curvature lens. In fact, having a curved lens with a planar image plane can be shown to be equivalent to the scenario of having a planar lens with a curved image plane; thus, the latter approach was adopted since it is similar to the human vision system illustrated by Netter (Figure 1). The curved image plane will be approximated as a sphere, therefore allowing spherical constraints to be adopted. The benefit of using spherical constraints, apart from its mathematical simplicity, is the absence of singularities with respect to incidence angles. Unlike Schneider et al. (2009), no special considerations are required for an incidence angle of 90°, and unlike Scaramuzza et al. (2006a), it can capture incidence angles that approach the limit of 180° (i.e. 360° FOV).

The 3D object space coordinates of the signalized targets are related to the camera coordinate system via a rigid body transformation (Equation 1).

$$V_{ij} = q_j (P_i - T_j) q_j^c \qquad (1)$$

where, $V_{ij} = [X_{cij}, Y_{cij}, Z_{cij}]^T$ is the rotated and translated object space coordinates of target $i$ in exposure $j$;
$P_i = [X_i, Y_i, Z_i]^T$ is the object space coordinates of target $i$;
$T_j = [X_{oj}, Y_{oj}, Z_{oj}]^T$ is the position of the camera in exposure $j$;
$q_j$ = unit quaternion representing orientation of the camera in exposure $j$. Superscript '$c$' represents the quaternion conjugate.

The incidence angle ($α_{ij}$) of the incoming light rays relative to the optical axis can then be computed from $V_{ij}$ using the tangent function (Equation 2). If the collinearity condition is to be satisfied, the angle of refraction ($β_{ij}$), computed from the corrected image coordinates $p_{ij} = [x_{ij} – x_p – Δx, y_{ij} – y_p – Δy, -c]^T$ (Equation 3), should be equal to the angle of incidence (i.e. $α = β$).

$$\tan(α_{ij}) = \frac{\sqrt{X_{cij}^2 + Y_{cij}^2}}{Z_{cij}} \qquad (2)$$

$$\tan(β_{ij}) = \frac{\sqrt{(x_{ij}^{true})^2 + (y_{ij}^{true})^2}}{\text{sgn}(Z_{cij}) z_{ij}} \qquad (3)$$

where, $x_{ij}^{true} = x_{ij} - x_p - Δx$ is the $x$ image measurement corrected for the principal point offset and distortion;
$y_{ij}^{true} = y_{ij} - y_p - Δy$ is the $y$ image measurement corrected for the principal point offset and distortion;
$z_{ij}$ is the $z$ coordinate of the homologous image measurement on the virtual curved image plane. It is assumed to lie on the surface of a sphere and obey the condition $(x_{ij}^{true})^2 + (y_{ij}^{true})^2 + (z_{ij})^2 = c^2$;
$\text{sgn}(Z_{cij})$ is either + or - depending if the target point is in front or behind the camera

Equating the two equations (i.e. Equations 2 and 3) and constraining the image points to lie on a sphere yields the following constraint (Equation 4). From this, the Karush-Kuhn-Tucker (KKT) conditions can be derived. Numerically, the KKT system of equations are solved iteratively as an equality constrained weighted implicit least squares adjustment. The update steps are calculated using the dogleg strategy, which is a trust-region method (Nocedal and Wright, 2006).

$$\frac{\sqrt{X_{cij}^2 + Y_{cij}^2}}{Z_{cij}} - \frac{\sqrt{(x_{ij}^{true})^2 + (y_{ij}^{true})^2}}{\text{sgn}(Z_{cij})\sqrt{c^2 - (x_{ij}^{true})^2 - (y_{ij}^{true})^2}} = 0 \qquad (4)$$

The corrections to the image measurements (i.e. $Δx$ and $Δy$) are estimated using k-nearest neighbour (kNN) regression on the bundle adjustment residuals in a grey-box system identification framework. Once the corrections are determined, the least squares adjustment is repeated to minimize the reprojection error and a new kNN regressor is trained using a 10-fold cross-validation approach. This process continues until a solution is reached where the image reprojection error and the weighted kNN training score are both stable and minimized. In contrast to previous publications, the error model (Equation 5) from Brown (1971) was not adopted. This is motivated by the fact that a perfectly spherical image plane is merely an assumption that is expected to be violated (e.g. some catadioptric camera systems use hyperbolic mirrors to achieve the wide FOV). Therefore, $Δx$ and $Δy$ are not only correcting for RLD, DLD, affinity, and shear, but also other lens/mirror misalignment, moving entrance pupil, the real mapping geometry for the lens and camera combination (which is assumed to be unknown from the user's perspective), and any other empirical errors of the sensor. In fact, some fisheye lenses are known to not hold the single viewpoint property perfectly and such a deviation from a central system can be modelled in $Δx$ and $Δy$ using a non-parametric method like kNN regression.

Since automatic target extraction and labelling was utilized in this project, some blunders are expected to exist in the image correspondences. Therefore, the iterative re-weighted least-squares with a Huber weight function is applied to ensure that the effect of outliers on the overall solution is reduced. Other robust M-Estimators can also be used, but for the data presented in this paper, the Huber function was empirically found to have the best balance between accuracy and robustness.

$$\begin{aligned}Δx &= \bar{x}(K_1 r^2 + K_2 r^4 + K_3 r^6) + P_1(r^2 + 2\bar{x}^2) \\ &\quad + 2P_2(\bar{x}\bar{y}) + A_1 \bar{x} + A_2 \bar{y} \\ Δy &= \bar{y}(K_1 r^2 + K_2 r^4 + K_3 r^6) + 2P_1(\bar{x}\bar{y}) + P_2(r^2 + 2\bar{y}^2)\end{aligned} \qquad (5)$$

where, $\bar{x} = x - x_p$;
$\bar{y} = y - y_p$;
$r = \sqrt{\bar{x}^2 + \bar{y}^2}$;
$K_1, K_2, K_3$ are coefficients for radial lens distortion;
$P_1, P_2$ are coefficients for decentering lens distortion;
$A_1, A_2$ represent affinity and shear (respectively)

## 4. EXPERIMENTATION

Coded targets of various sizes were uniformly distributed on the ceiling, floor, and walls of a room. 44 digital photos were captured using a Nikon D600 DSLR with a 28 mm focal length (i.e. ~75° diagonal FOV) and 52 photos were captured using a GoPro Hero3 Silver Edition camera with a 17 mm focal length (i.e. ~150° diagonal FOV) by means of an "inside-out" approach. Sample pictures acquired using both sensors are shown in Figures 2 and 3. The centroids of all the targets in the images were measured and labelled automatically using the software Photomodeler. Some coded targets required the user to indicate a bounding-box before the automatic target extraction worked properly. However, since there was an abundance of coded targets and many photos were captured, even if some targets failed to be detected there was still a sufficient amount of redundancy.

Since the presented calibration method relies on a non-linear estimator, a reasonable initialization of the unknown camera pose and object space target positions is required. The approximate centroid coordinates of the targets are obtained from surveying to establish a scale for the photogrammetric network, and the initial approximation of the camera poses is determined using the perspective-three-point algorithm described in Gao et al. (2003). This is preferred over the direct linear transformation approach because the over-parametrization requires targets that occupy a 3D volume, which was not satisfied by many of the images captured by the Nikon DSLR.

In order to assess the 3D reconstruction accuracy of the photogrammetry solution, 191 targets were measured using a FARO Focus 3D terrestrial laser scanner (Figure 4). High density 360° scans of the 3D target field were acquired from two stations at opposite corners of the room. A least-squares 3D geometric form fitting was performed to determine the centroid of each paper target (Chow et al., 2010). The surveyed results from the two independent stations were then registered together using a 3D rigid body transformation and the final target centroids were statistically inferred from the weighted average of the spatial distance and direction measurements, post-registration (Lichti, 2007).

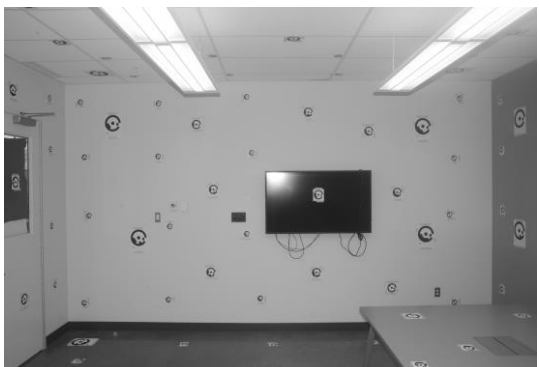

Figure 2: Sample photo captured by the Nikon DSLR

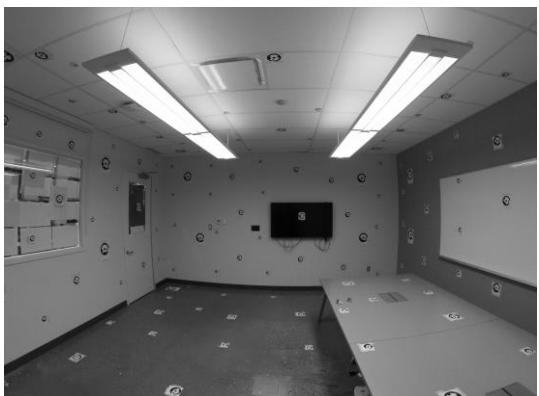

Figure 3: Sample photo captured by the GoPro

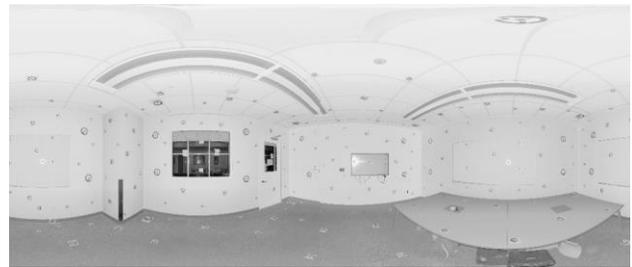

Figure 4: Panoramic image view of the laser scanner data

## 5. RESULTS AND ANALYSIS

### 5.1 Nikon D600 DSLR: ~75° FOV

The convergence of the proposed calibration method is shown in Figure 5. A stable minimum for the cost function could be found after about 50 iterations and the final a posteriori variance factor is slightly below one. This a posteriori variance factor is lower than the conventional adjustment with only $K_1$ but remains higher than the scenario where $K_1$ and $K_2$ are modelled.

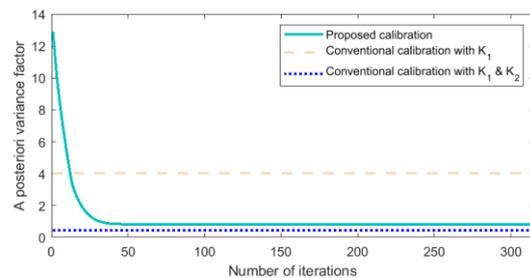

Figure 5: The minimization of cost after each iteration

The geometric reconstruction error of the proposed method was also compared to the conventional bundle adjustment with self-calibration approach. The calibration was performed under a highly redundant network (i.e. with a redundancy of 3383) to allow for inference of the AP. The number of AP in the conventional method was chosen using a combination of graphical analysis techniques and statistical testing. In the conventional model, the best correction model that balances both bias and variance is when the AP consist of only $K_1$ and $K_2$. Table 1 summarizes the image space error (i.e. the least-squares cost, which is the summation of the weighted squares of the reprojection error) and the root mean square error (RMSE) of the triangulated 3D mapping space when compared to the adjusted terrestrial laser scanning survey. The results obtained when estimating the XYZ object space, EOP, IOP, and AP using kNN regression showed a similar level of error as the conventional approach. While the proposed approach was able to achieve a better mapping accuracy than the conventional approach with only $K_1$, if both $K_1$ and $K_2$ are estimated, the standard approach to camera calibration outperforms the proposed method.

| Model | Image Space Error | RMSE [mm] | | |
|---|---|---|---|---|
| | | X | Y | Z |
| XYZ, EOP, IOP | 4.4E+04 | 16.9 | 17.4 | 10.3 |
| XYZ, EOP, IOP, $K_1$ | 1.4E+04 | 2.9 | 2.4 | 1.5 |
| XYZ, EOP, IOP, $K_1$, $K_2$ | 1.5E+03 | 1.1 | 1.0 | 0.7 |
| XYZ, EOP, IOP, $K_1$, $K_2$, $K_3$ | 1.4E+03 | 1.1 | 1.0 | 0.7 |
| XYZ, EOP, IOP, $K_1$, $K_2$, $K_3$, $P_1$, $P_2$ | 1.2E+03 | 1.1 | 1.0 | 0.7 |
| XYZ, EOP, IOP, $K_1$, $K_2$, $K_3$, $P_1$, $P_2$, $A_1$ | 1.1E+03 | 1.1 | 0.9 | 0.6 |
| XYZ, EOP, IOP, $K_1$, $K_2$, $K_3$, $P_1$, $P_2$, $A_1$, $A_2$ | 1.1E+03 | 1.1 | 0.9 | 0.6 |
| XYZ, EOP, IOP, kNN | 2.7E+03 | 1.5 | 1.2 | 0.7 |

Table 1: Quality assessment of the Nikon DSLR calibration

Overall, the post-calibration dispersion of residuals decreases after using both calibration methods. However, the conventional calibration results in higher precision in both x and y directions (Figure 6). Figure 7 shows the histogram of the image residuals from both calibrations. Although the standard deviation of the residuals is lower in the conventional bundle adjustment, the Gaussian curve is less peaked due to mild asymmetry in both the positive x and y directions (Figure 6).

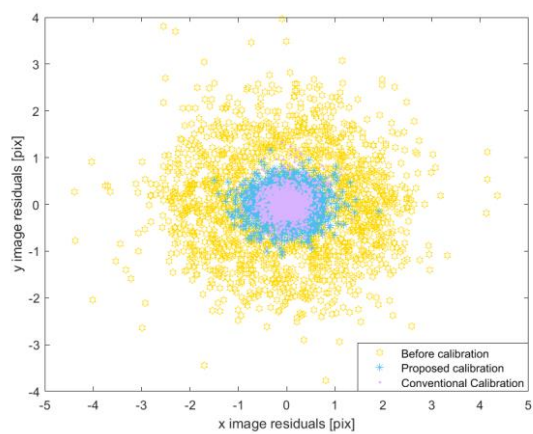

Figure 6: 2D distribution of the Nikon DSLR image residuals

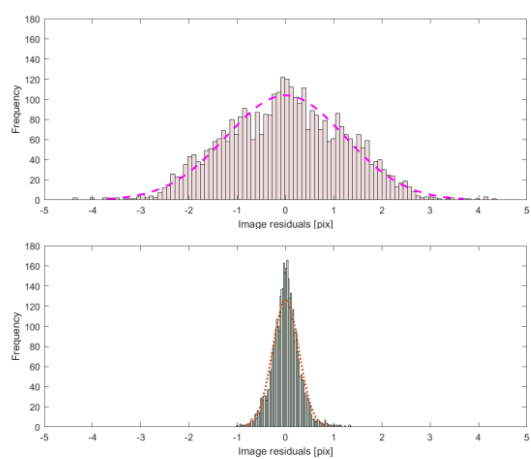

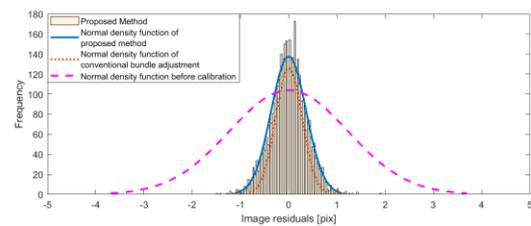

Figure 7: Histogram of Nikon DSLR image residuals before calibration (top), after conventional calibration (middle), and after proposed calibration (bottom)

### 5.2 GoPro Hero3 Silver Edition: ~150° FOV

Through testing different combinations of systematic error terms, it was found that the best camera correction model (using the conventional bundle adjustment with a redundancy of 3369) included $K_1$, $K_2$, $K_3$, $P_1$, and $P_2$. This was expected for a wider-angle lens, as high-order correction terms are statistically significant and important in compensating for the RLD. Table 2 summarizes the quality control results of a few different bundle adjustment models. The results using the proposed kNN regression model appear in the last row and have the smallest reprojection and mapping errors.

| Model | Image Space Error | RMSE [mm] | | |
|---|---|---|---|---|
| | | X | Y | Z |
| XYZ, EOP, IOP | 5.5E+05 | 62.0 | 68.1 | 69.5 |
| XYZ, EOP, IOP, $K_1$ | 1.2E+05 | 16.1 | 14.0 | 13.9 |
| XYZ, EOP, IOP, $K_1$, $K_2$ | 3.1E+04 | 7.7 | 6.6 | 6.2 |
| XYZ, EOP, IOP, $K_1$, $K_2$, $K_3$ | 1.4E+04 | 6.8 | 6.3 | 4.8 |
| XYZ, EOP, IOP, $K_1$, $K_2$, $K_3$, $P_1$, $P_2$ | 1.0E+04 | 3.9 | 3.3 | 2.7 |
| XYZ, EOP, IOP, $K_1$, $K_2$, $K_3$, $P_1$, $P_2$, $A_1$ | 1.0E+04 | 3.9 | 3.3 | 2.7 |
| XYZ, EOP, IOP, $K_1$, $K_2$, $K_3$, $P_1$, $P_2$, $A_1$, $A_2$ | 1.0E+04 | 3.9 | 3.3 | 2.7 |
| XYZ, EOP, IOP, kNN | 4.6E+03 | 1.6 | 1.7 | 1.3 |

Table 2: Quality assessment of the GoPro Hero3 calibration

The x and y image residuals can be visualized in Figure 9. Both solutions are quite precise; however, the residuals after the proposed calibration (where the distortion profile was learned using kNN regression) exhibit even less dispersion than the conventional bundle adjustment approach. Figures 10 and 11 show the distribution of residuals using the two bundle adjustment methods. Not only is the distribution narrower and more peaked when using the proposed method, a slight skew can be observed in the distribution of y image residuals from the conventional approach. Upon further investigation of the conventional solution, it was found that a linear trend remains in the y image residuals post-calibration (Figure 12). This linear trend in the conventional method (i.e. slope = -1.7 x $10^{-4}$ and intercept = 0.21) has been adequately modelled by the kNN regressor (i.e. slope = -3.5 x $10^{-5}$ and intercept of 0.04) and is essentially unobservable in the proposed approach.

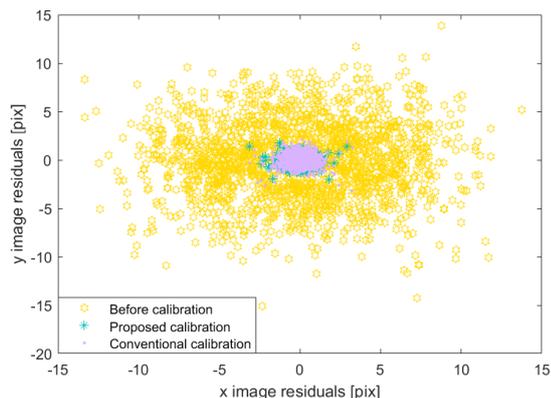

Figure 9: 2D distribution of the GoPro Hero3 image residuals

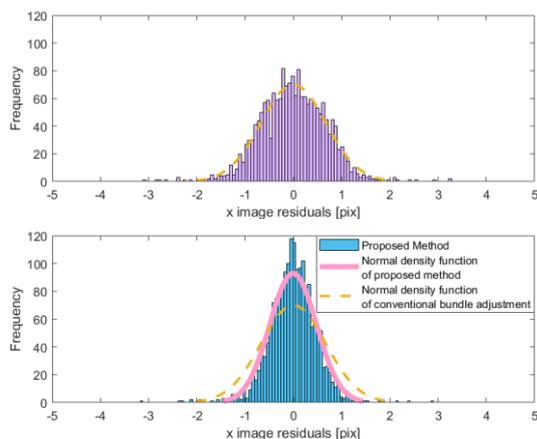

Figure 10: Histogram of GoPro Hero3 x image residuals after conventional calibration (top) and after proposed calibration (bottom)

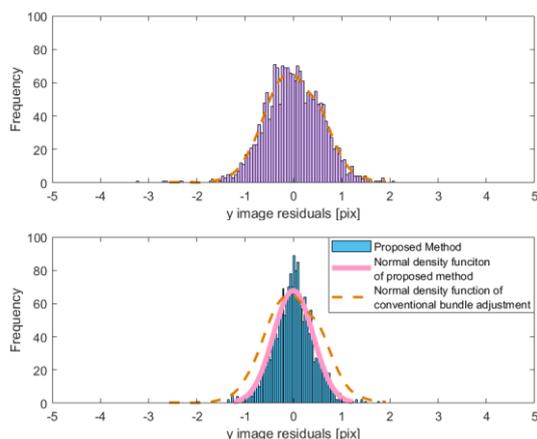

Figure 11: Histogram of GoPro Hero3 y image residuals after conventional calibration (top) and after proposed calibration (bottom)

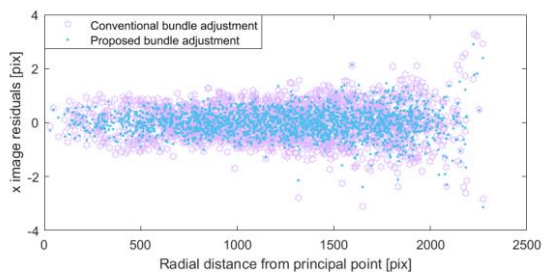

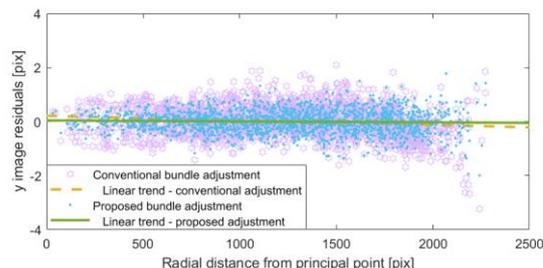

Figure 12: Image residuals of GoPro as a function of radial distance

## 6. DISCUSSION AND CONCLUSION

With the known benefits of a wide FOV camera for navigation, companies are expected to pursue technology breakthroughs to satisfy the market pull by designing camera systems with FOVs that approach a full sphere. For example, the Entaniya Fisheye 250 lens that can look behind the sensor itself with a FOV of 250° was released in 2016. In this paper, a "nearly" fully-automatic, end-to-end wide-angle calibration method general to any catadioptric or dioptric camera systems is presented. It is "nearly" because the target extraction and labelling software (Photomodeler) would regularly fail to identify targets, thereby requiring some manual intervention to provide a search area to scan for them. This issue can be resolved, for instance by replacing the coded targets with a planar checkerboard target field and using the automatic corner extraction method in the Omnidirectional Camera Calibration Toolbox for MATLAB (Scaramuzza et al., 2006b).

The proposed method has the benefits of not requiring the following:

- prior knowledge about the imaging sensor,
- specialized equipment (e.g. robotic arms or turntables),
- visibility of the circular image boundary, or
- a human operator to choose the distortion model

The user is free to place coded targets anywhere, provided that approximate coordinates of the targets are known, and photos of the targets are captured in both landscape and portrait orientations from different positions. The initial approximation of the camera poses is determined using the perspective-three-point algorithm. This is subsequently refined by performing a free-network photogrammetric adjustment with an equality constraint on the incidence and refraction angles to maintain collinearity. By formulating the bundle adjustment with image coordinates lying on the surface of a sphere, self-calibration can be performed without restriction on the incidence angle, thus allowing omnidirectional cameras with any FOV to be modelled (unlike previous research).

Two cameras (one with a normal FOV lens and one with a wider FOV lens) were calibrated. When the proposed method was used for calibrating a normal lens camera, the 3D measurement error was reduced from centimetre-level to millimetre-level. This result is comparable to a well-established camera calibration method, with an overall difference in 3D reconstruction accuracy of 0.4 mm or less.

More importantly, the proposed method performed better than the conventional method when dealing with wider-angle lens cameras. The conventional bundle adjustment approach was unable to completely alleviate the systematic distortions in the camera system, whereas the proposed method was able to handle the high degree of distortions better. When compared to the conventional approach, using the proposed method yielded an average of 53% improvement in 3D mapping accuracy.